 \documentclass{vgtc}                          

\ifpdf
  \pdfoutput=1\relax                   
  \usepackage{graphicx}                
  \DeclareGraphicsExtensions{.pdf,.png,.jpg,.jpeg} 
\else
  \usepackage{graphicx}                
  \DeclareGraphicsExtensions{.eps}     
\fi%

\graphicspath{{figures/}{pictures/}{images/}{./}} 

\usepackage{microtype}                 
\PassOptionsToPackage{warn}{textcomp}  
\usepackage{textcomp}                  
\usepackage{makecell}
\usepackage{mathptmx}                  
\usepackage{times}                     
\usepackage{cite}                      
\usepackage{tabu}                      
\usepackage{booktabs}                  
\usepackage{color}

\onlineid{0}

\vgtccategory{Research}

\vgtcinsertpkg

\title{airpred: A Flexible R Package Implementing Methods for Predicting Air Pollution}

\author{M. Benjamin Sabath, MA \thanks{e-mail: mbsabath@hsph.harvard.edu} %
\and Qian Di, ScD \thanks{e-mail: qiandi@mail.harvard.edu}%
\and Danielle Braun, PhD \thanks{e-mail: dbraun@mail.harvard.edu}
\and Francesca Dominici, PhD \thanks{email: fdominic@hsph.harvard.edu}
\and Christine Choirat, PhD \thanks{email: cchoirat@iq.harvard.edu}
}
\affiliation{\scriptsize Department of Biostatistics, Harvard T. H. Chan School of Public Health}

\abstract{
Fine particulate matter (PM$_{2.5}$) is one of the criteria air pollutants regulated by the Environmental Protection Agency in the United States. There is strong evidence that ambient exposure to (PM$_{2.5}$) increases risk of mortality and hospitalization. Large scale epidemiological studies on the health effects of PM$_{2.5}$ provide the necessary evidence base for lowering the safety standards and inform regulatory policy. However, ambient monitors of PM$_{2.5}$ (as well as monitors for other pollutants) are sparsely located across the U.S., and therefore studies based only on the levels of PM$_{2.5}$ measured from the monitors would inevitably exclude large amounts of the population. One approach to resolving this issue has been developing models to predict local PM$_{2.5}$, NO$_2$, and ozone based on satellite, meteorological, and land use data. This process typically relies developing a prediction model that relies on large amounts of input data and is highly computationally intensive to predict levels of air pollution in unmonitored areas. We have developed a flexible R package that allows for environmental health researchers to design and train spatio-temporal models capable of predicting multiple pollutants, including PM$_{2.5}$. We utilize H2O, an open source big data platform, to achieve both performance and scalability when used in conjunction with cloud or cluster computing systems.
} 

\CCScatlist{ Machine Learning, Air Pollution Prediction, Environmental Health 
}

\begin{document}



\maketitle

\section{Introduction}
\subsection{PM$_{2.5}$ and Health}
Fine particulate matter (PM$_{2.5}$) is one of criteria air pollutants regulated by the Environmental Protection Agency in the United States. The main sources of (PM$_{2.5}$) are fossil fuel combustion [add reference. It is defined as suspended particles of solid or liquid less than 2.5 micrometers in diameter. 

There is strong evidence of an association between short and long term exposure to ambient levels  of PM$_{2.5}$ and mortality, hospitalization and many other adverse health outcomes. High levels of PM$_{2.5}$ are associated with decreased life expectancy \cite{beelen_effects_2014}. Additionally, studies have shown that even low levels of PM$_{2.5}$ are associated with increased mortality\cite{di_air_2017, di_association_2017}. The effect at low levels is key as Environmental Protection Agency (EPA) regulations currently set the legal limit for PM$_2.5$ at an annual average of 12 $\mu$g per cubic meter. Importantly, these studies found no evidence of a threshold effect, and noted that increased mortality associated with PM$_{2.5}$ exposure continued to be observed at levels down to 5 $\mu$g per cubic meter. 

\subsection{Challenges in Studying PM$_{2.5}$}

\begin{figure}
    \centering
    \includegraphics[width=0.5\textwidth]{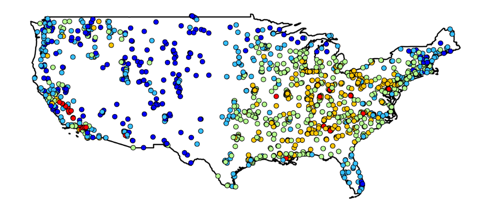}
    \includegraphics[width=0.5\textwidth]{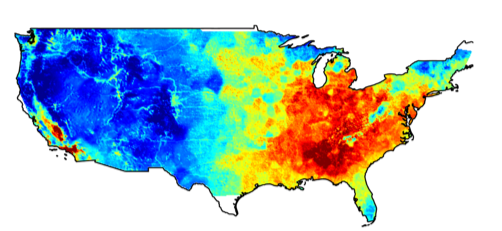}
    \caption{Distribution of pollution monitors and of predictions from Di el al. model}
    \label{fig:my_label}
\end{figure}

Studying the effects of PM$_{2.5}$ presents a number of challenges to researchers. Traditionally studies have relied on exposure data from PM$_{2.5}$ monitors\cite{di_air_2017}. This type of analysis has limitations; as health effects of PM are local, analysis relying on monitor data is limited to study populations in which monitors are located. PM$_{2.5}$ monitors tend to be placed in urban areas and other population centers, restricting studies to those areas. In addition, monitors are sparse and do not represent the entire population of interest. The typical study cohort for PM studies tends to be more urban and affluent than the actual population distribution \cite{di_air_2017}. 


In order to more accurately estimate the health effects of PM$_{2.5}$ exposure, multiple groups have attempted to develop methods of estimating PM$_{2.5}$ in locations where monitors are not present\cite{di_assessing_2016,van_donkelaar_high-resolution_2015}. This most typically takes the form of statistical and machine learning models using satellite and land use data to project values of PM$_{2.5}$ and other pollutants onto areas without sustained measurements.

\subsection{Modelling Approaches}

Various groups have developed air pollution modelling platforms, with the goal of using information that is available in locations with and without monitors to predict ground level PM$_{2.5}$. A key component used in many of the developed models is aerosol optical depth (AOD), a measure of visibility that is associated with levels of particulate matter in the atmosphere. However, these measurements represent particulate matter present in the entire atmospheric column, and can't provide a reliable proxy to ground level pollution in and of themselves\cite{van_donkelaar_high-resolution_2015}. Information from satellite, meteorological and land use sources is then used to attempt to estimate the levels of PM$_{2.5}$ at surface. 

One key form of data with universal coverage used to help model the levels of PM and other pollutants are the results of chemical transport models (CTMs). CTMs are computationally intensive atmospheric models that model material flows and chemical reactions within the atmosphere. When combined with measurements of AOD, these have been shown to provide more accurate measurements of PM$_{2.5}$\cite{van_donkelaar_high-resolution_2015,liu_mapping_2004}. Multiple approaches also bring in land use data to further refine the predictions \cite{di_assessing_2016,van_donkelaar_high-resolution_2015}. These pieces of information are combined either using standard linear based statistical models or more complex machine learning methods in order to generate predictions. These are typically able to generate predictions at a 1km x 1km scale for large regions, such as the continental United States\cite{di_air_2017} and the whole of North America\cite{van_donkelaar_high-resolution_2015}.

Two models of note that were developed using this paradigm are the model published by Di et al.\cite{di_assessing_2016} and that published by van Donkelaar et al.\cite{van_donkelaar_high-resolution_2015} Van Donkelaar's group takes the approach of using levels AOD predicted by the GEOS-Chem CTM as priors for an optimal estimation (OE) model that takes estimates of PM$_{2.5}$ from the Moderate Resolution Imaging Spectroradiometer (MODIS) satellite as inputs. A geographic weighted regression is then fit using information about urban land cover, elevation, and GEOS-Chem projections of PM$_{2.5}$ chemical components (the chemical composition of the particles that make up the total suspended particulate matter) to model the bias of the OE model compared to the measurements of PM from the AERONET sun photometers. One point of interest in this approach is that the researchers intentionally chose to not include information from ground based monitors as they wanted their method to be applicable in locations where few monitors were available, such as Northern Canada.

In contrast to this, the model developed by Di et al\cite{di_assessing_2016}, relies heavily on data from PM$_{2.5}$ monitors to generate predictions of PM$_{2.5}$. They take a data intensive approach, using information about AOD from the MODIS satellite, surface reflectance data, estimates of both ground level PM$_{2.5}$ and total levels of aerosols distributed throughout the whole atmospheric column from GEOS-Chem, meteorological data from the North American Regional Reanalysis project, indices of aerosols that could potentially absorb PM$_{2.5}$, and land use information such as elevation, road density, vegetation coverage, and population density that can serve as reasonable proxies for emissions as well as help capture small scale variations in PM levels.

Both of these models have been applied in health research to determine the effects of low level PM$_{2.5}$. In the case of the Di model, annual averages of PM$_{2.5}$ estimates were linked to zip codes, and then used in conjunction with data on Medicare all cause mortality to assess the risks of exposure to PM$_{2.5}$. By incorporating the predictions this research was able to analyze data from underrepresented populations and demonstrate a connection between increased mortality and PM exposure even at low levels\cite{di_association_2017,di_air_2017}. These results have also been confirmed by other studies using different PM prediction techniques in both Europe and Canada\cite{pinault_risk_2016,pinault_associations_2017,dehbi_air_2017}.

\subsection{Definitions of Terms}
There are a number of terms used throughout this paper that require a definition to place them in the context of our work. We define a training set as a dataset where the outcome variable is known, and all covariates are known or imputed, while a test set or an additional set of interest is defined as a set where all covariates are known or imputed, but the outcome variable is unknown. We define identifying variables as the set of variables that are used to uniquely identify an observation, akin to the concept of the key from database design.

\subsection{Challenges in Modelling PM$_{2.5}$}

A unifying factor among all approaches is that they are both data and computationally intensive. In our work, we use data that take up 30 TBs of disk storage in their unprocessed form. Therefore all research on methods of modeling PM$_{2.5}$ and other pollutants must take into account technical limitations at every step of the process. Moving large quantities of data through memory, let alone feeding them into neural networks or other modeling methods can quickly run into system limits.
Due to this, it is important to take into account not just the methodology of prediction, but the entire work flow, and to treat all work around it as a single prediction platform.

As a solution to this issue, we have developed an R package that we are calling "airpred" which implements a single prediction platform for modeling air pollution exposure data. We begin by processing raw data so that it can be used for prediction. We then develop the prediction platform, which has two components. The first component includes model training using methods that maintain their performance advantage at a higher spatial resolution over more intensive methods (i.e. CTMs). 
The second includes obtaining model predictions, using the trained models to generate reasonable predictions from additional data sets of interest. 
Although the methods described in the following section use PM$_{2.5}$ as an example, the developed R package is flexible and can be applied to any pollutant.

\section{Problem Definition}


There are a number of elements that we needed to consider when designing our platform. First of which is what types of data sources would we be using, both in terms of file type and information represented. Secondly, what features would users need, and what would be the best way for users to interact with our code. Finally, what would the computational infrastructure for our platform look like, what would be the computational demands, and what would the most computationally efficient way of processing the data be.

\subsection{Data Sources}

The inputs for the PM$_{2.5}$ model come from a variety of sources including atmospheric imaging from primarily NASA satellites, meteorological results of CTM simulations, geographical information, and information on land usage (such as measurements of road density). These data are downloaded directly from their respective sources or are acquired from public databases and thus have varying file types and spatial resolutions. 

\begin{table}[]
    \centering
    \caption{Data Sources and Resolutions used in the Di Model}
    \begin{tabular}{c||c|c}
        Input Type & Sources & \makecell{Spatial \\Resolution} \\ \hline\hline
        Meteorological Data & Reanalysis & 0.5\textdegree x 0.625\textdegree \\ \hline
        AOD & MAIACUS & 1km x 1km \\ \hline
        Surface Reflectance & MOD09A1 & 500m x 500m \\ \hline
        CTMs & \makecell{GEOS-Chem\\ CMAQ} & 12k x 12k \\ \hline
        Absorbing Aerosols & \makecell{OMAERUVd\\ OMAEROe} & 0.25\textdegree x 0.25\textdegree \\ \hline
        Vegetation & MOD13A2 & 1km x 1km \\ \hline
        Other Land Use & NLCD & 30m x 30m

    \end{tabular}
    \label{tab:my_label}
\end{table}

We designed the package to take in the data sources used by the Di model. These data are all available from public sources. However, all these data come in multiple file types and have different spatial resolutions. The data include rasters, centroids, text files, and shape files with resolutions spanning from 30m x 30m for the land use data to 12km x 12km for the CTM inputs (Table 1).

After downloading the raw data, inverse distance weighted interpolation is performed to assign values to locations of interest, which are either the location of monitors that are being used in the training set, or are other regularly distributed locations being used for prediction (commonly called grid points) in other data sets of interest. Due to the variation in spatial and temporal resolution, there is some missingness generated during the interpolation process. Some sources, notably those measuring surface reflectance, only have observations for a given location on a biweekly basis. In order to be assembled, the data need to be arranged so that there is a single spatial resolution, with a regular temporal resolution. In practice, we use daily data for temporal resolution, and either the locations of the monitors or a 1km x 1km grid as a spatial reference. The 1km x 1km grid was selected because this is the spatial resolution of the AOD data. 

We developed code as part of the R package to assemble the interpolated data into a data frame like structure where each row represents the data for a given site (either a monitor or a prediction grid location) on a given day. 
In order to further prepare the data, a series of normalizations (a rescaling of variables to be between zero and one), imputations (using information in our dataset to estimate the values of missing data that are not the outcome variable), and transformations (rescaling the values of a variable so that outliers are closer to the mean) are also performed.

We do each of these steps for a number of reasons. We implement the imputation step in order to increase the amount of information available during model training, as without this process, observations with missing data would be excluded from model training. We perform normalization as most machine learning algorithms are designed assuming that all features are scaled between zero and one. Finally, we have the option of transformation as the imputation method implemented in the most recent implementation of the model described by Qian Di et. al\cite{di_assessing_2016} requires a transformation step to function well.  

\subsection{User Needs}
The target audience of the package primarily consists environmental epidemiology researchers and environmental scientists.
As R\cite{r_core_team_r:_2017} is a common programming language used in these two fields, we chose to implement the processing in the form of an R package. In addition to user familiarity, R also has the advantage of having a large ecosystem of packages available allowing for us to use systems developed by others rather than needing to develop many utilities ourselves. Additionally, by developing in R, we are able to potentially release our package on CRAN, a repository of publicly available R packages that can be accessed via built in functions in R. All R packages are open source as well, allowing for us to ensure that all code used is publicly available.

Additionally, we wanted to develop a platform that allows for flexibility, as both the model inputs and the statistical models themselves are are frequently changing. 
Given these frequent changes, we felt it was important to design a system that would allow users to easily change these elements without making any changes to the back-end code of the package. 

Further, we felt it was important to ensure readability and ease of use for any script utilizing the developed package. Therefore we endeavored to minimize the number of arguments passed directly to functions, as well as designed the package to include only a small number of clearly named functions that users would need to call in order to implement the full work flow. 

\subsection{Computational Limitations}

As a language, R does not lend itself well to solving big data problems. It holds all objects in memory and  parallel computing is not implemented as a default in the system. Minimizing memory impact and taking advantage of parallel algorithm structure are some of the main strategies for dealing with the issues that analyzing large data sets (such as the one used to predict pollution) present. Given this, any code written in R that did not rely on other systems was not likely to give us the required performance. Solutions to this problem include selecting a system to wrap that improve memory management and parallelism, as well as considering hardware based solutions, such as leveraging a graphics processing unit (GPU).

\section{Implementation of the R Package}

In this section we address the processes that are implemented within the package, as well as documenting the systems that the R package relies on. We also illustrate the structures that we developed to allow the package to behave flexibly, as well as examining at a high level the structures that users interact with.


\subsection{Use of H2O}

\begin{figure}[]
    \centering
    \includegraphics[width=0.4\textwidth]{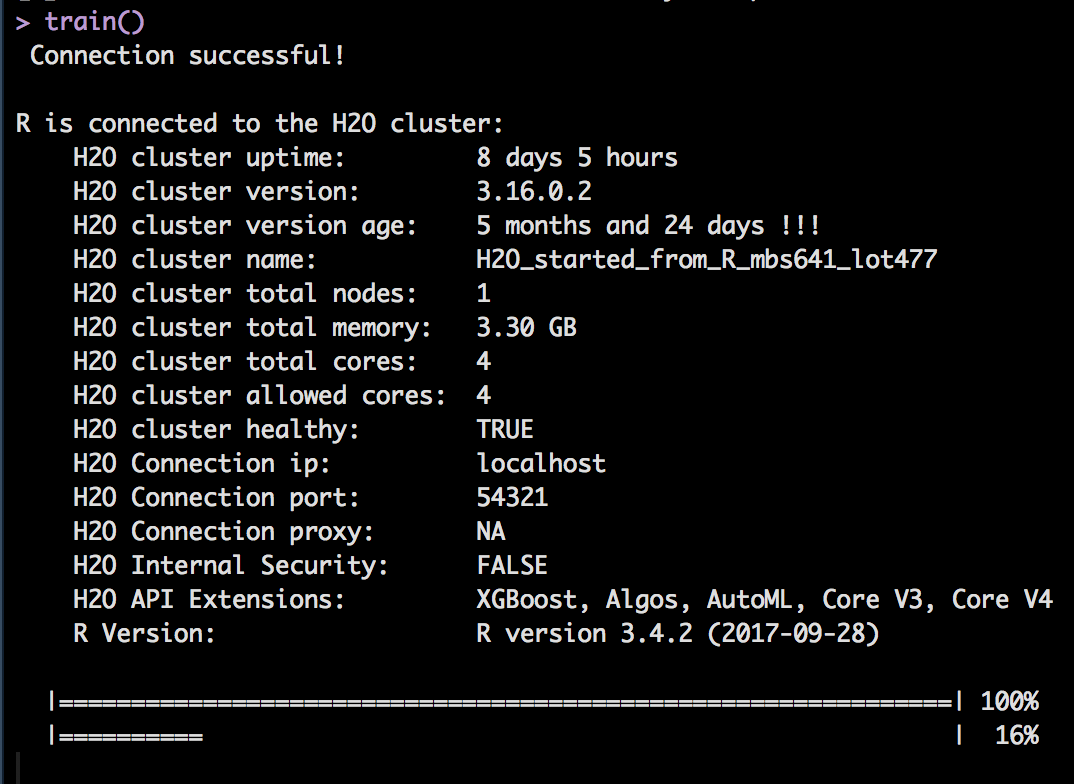}
    \includegraphics[width=0.4\textwidth]{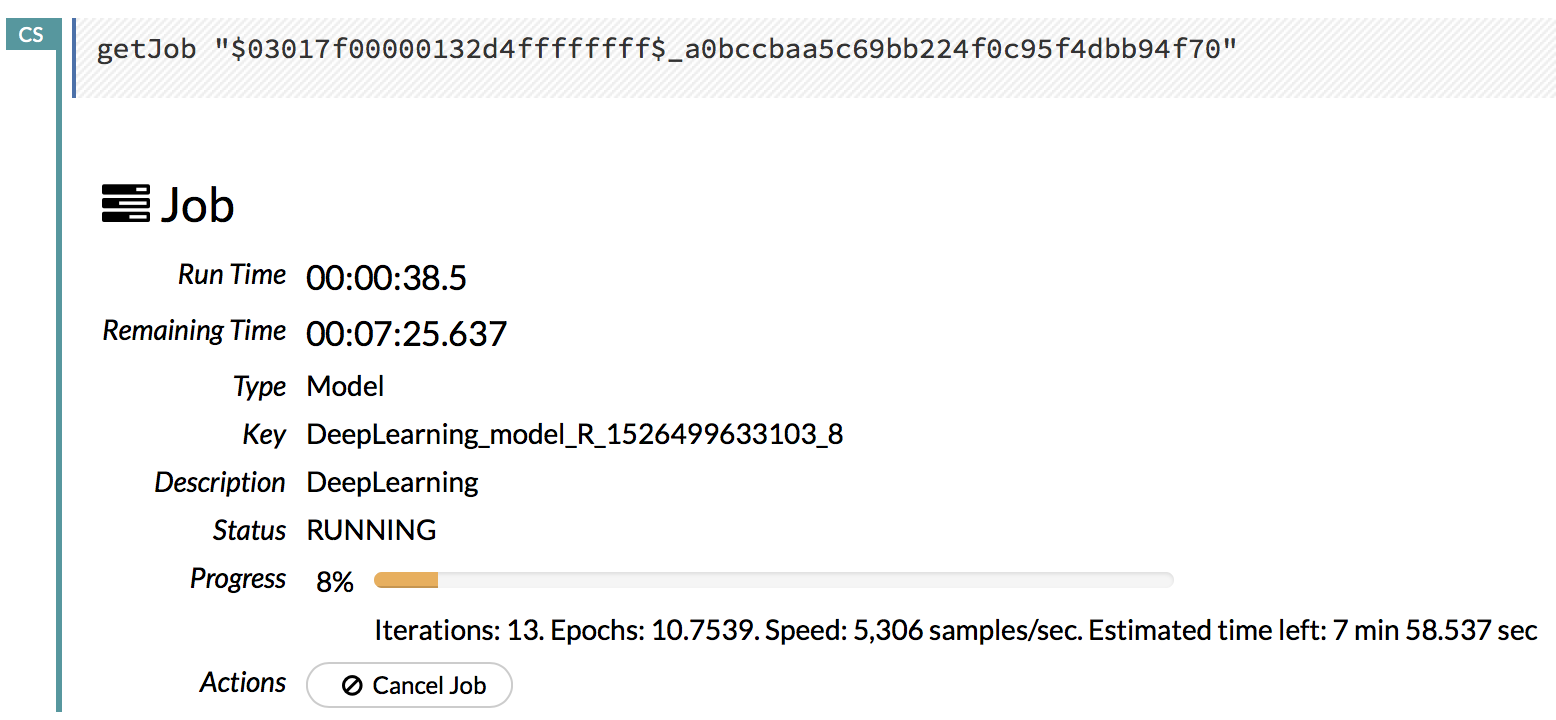}
    \caption{The H2O interface in R compared to the native H2O interface}
    \label{fig:my_label}
\end{figure}

H2O\cite{the_h2o_ai_team_h2o:_2017} is a powerful, open source system designed to run complex machine learning algorithms on large data sets. It performs well on large computer systems as well as laptops, facilitating the use of parallel algorithms to speed processing. The training steps of the developed airpred R package lean heavily on a number of features of H2O, while automating the initialization of the H2O cluster (if desired) as well as the direct passing of calls to the H2O cluster, instead allowing the user to specify all parameters of interest in configuration files.

In order to use H2O, first an H2O cluster must be initialized on the system that will be running the calculations. Despite sharing a name, H2O clusters bear little relation in function to cluster computing systems. Cluster computing systems consist of a series of servers, typically having a large number of CPUs and a high volume of memory. Users request resources to run computing jobs, and the resources are granted by an automated job scheduler if they are not in use by a different job. In contrast, H2O clusters are a set of CPUs, GPUs, and memory that are managed by the H2O system. A user sends an instruction to the H2O system, which then determines the most efficient way to utilize the resources available to it to complete the job (typically either training a machine learning model or getting predictions out of an already trained model). H2O functions well when used to with highly parallel algorithms and data that is large, but still able to fit in memory on research computing systems.

H2O has an R interface that airpred wraps, as well as a browser based interface to allow users to monitor the status and resource utilization of the cluster. All functions output progress bars to the R console, allowing users to track and troubleshoot the model training in real time. 


\subsection{Other Dependencies}

We incorporate a number of other packages into airpred in order to minimize the amount of new development we had to do, as well as to increase the maintanability of the code. We used data.table\cite{dowle_data.table:_2017} to handle joining of data, as well as to help read comma separated value (CSV) files directly from disk. We used the RANN\cite{arya_rann:_2017} package to implement the nearest neighbors based spatial interpolation algorithm used by Di et al.\cite{di_assessing_2016} in their model. We used the YAML\cite{stephens_yaml:_2016} package to help process our configuration files. We use the dplyr\cite{wickham_dplyr:_2017} and the lubridate\cite{grolemund_dates_2011} packages to help reshape data and process dates respectively. We use lme4\cite{bates_fitting_2015} to make linear mixed effect models available for use during the imputation step. R.matlab\cite{bengtsson_r.matlab:_2016} is used to read in matlab matrices during dataset assembly. Finally, we use the bam model from the mgcv\cite{wood_fast_2011} package to implement a parallalized GAM model.

\subsection{Algorithms}

\begin{figure}
    \centering
    \includegraphics[width=0.5\textwidth]{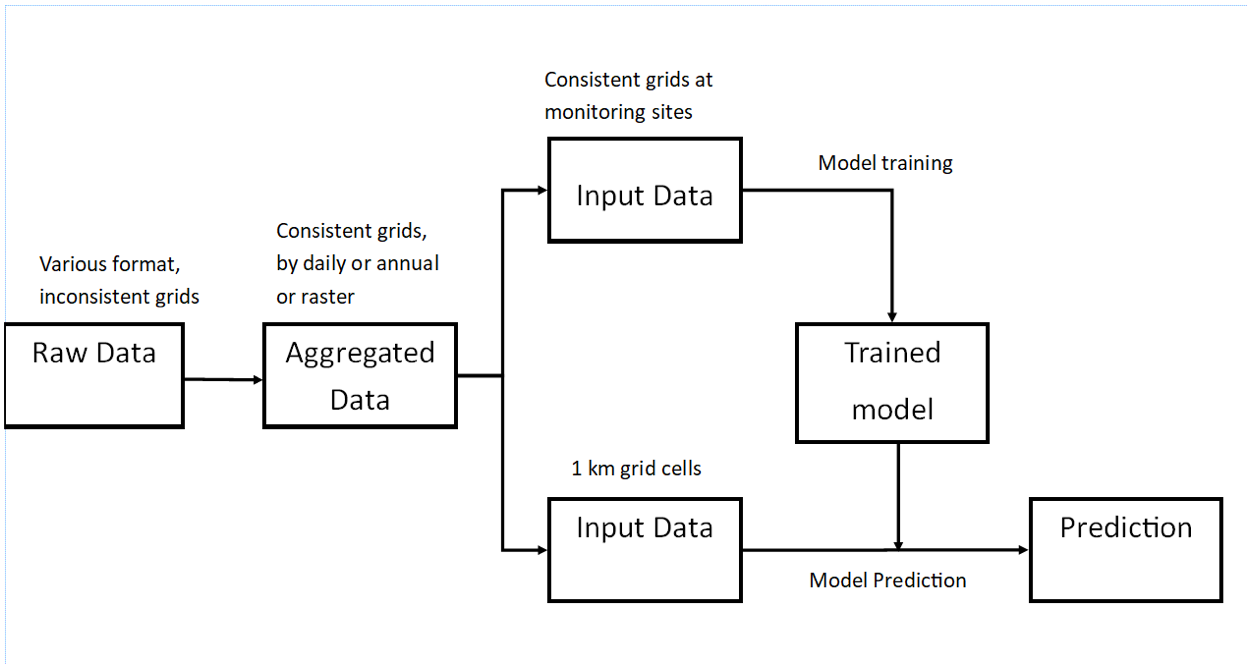}
    \caption{Work flow of using airpred package}
    \label{fig:my_label}
\end{figure}

This section covers the main processing steps within the airpred package. A brief summary of the flow of information within the package is as follows: Data is read into memory from disk, potentially assembled from binary matrix files, and has a number of pre-processing steps run on it. This cleaned data is used as a training set for a user selected single machine learning models or series of machine learning models combined in an ensemble model. 

\emph{Data Reading}
The raw data inputs used in our air pollution modelling work come from a variety of sources, have differing temporal and spatial resolutions, as well as differing file types. Given this, we felt it was best to not directly incorporate the initial data cleaning into the package, as every source would likely require its own subroutines. Instead, we assume that some initial cleaning has been done, converting the data into either a tree of matrices stored on disk representing the values of different data at different times or a data frame like structure that can be quickly read into R. While this forces some work to be done outside of the package work-flow, it allows the code to be flexible with regard to the problems it attempts to solve.


\emph{Data Processing}
Once the data is read and is arranged such that each observation in the dataset represents a measurement of all variables at a given site at a given point in time (Table 2) 
we then all the users the option to perform a series of imputations, normalizations, and transformations on the data. These, if chosen to be used, fill in missing data, re-scale values to exist between zero and one, and re-scale the distribution of values to lessen the impact of outliers respectively. The use of these is controlled in the primary configuration file. The data is saved after each step in the preparation process in order to create a record of changes for documentation, reproducibility, and troubleshooting purposes.

\begin{table}
    \centering
    \caption{A hypothetical example of the organization of data ready to be used for training with airpred. Site No. and Date are the key variables in this example.}
    \begin{tabular}[width = 0.5\textwidth]{c|c|c|c|c|c}
        Site No. & Date & MonitorData & Var1 & Var2 & ... \\ \hline\hline
         1 & 01/01/2000 & 5.5 & Value1 & Value2 & ... \\ \hline
         2 & 01/01/2000 & 5.6 & Value3 & Value4 & ... \\ \hline
         3 & 01/01/2000 & 5.4 & Value5 & Value6 & ... \\ \hline
         ... & 01/01/2000 & ... & ... & ... & ... \\ \hline
         1 & 01/02/2000 & 5.7 & Value7 & Value8 & ... \\ \hline
         ... & ... & ... & ... & ... & ... \\ \hline
         
    \end{tabular}

    \label{tab:my_label}
\end{table}

\emph{Model Training}
After the data is processed, it is passed to the model training algorithm. The airpred package allows for users to specify a set of machine learning models to be trained as part of the modelling process. The current implemented models are a neural net, random forest, and gradient boosting models. The training algorithm uses a general additive model (GAM) 
 model to determine weights for an ensemble of models specified by the user. 
 GAM's fit a model assuming that the variable in question is a sum of weighted smooth univariate functions of covariates\cite{hastie_generalized_1999}. The model assumes that all non-key variables 
 are inputs for the model being run. The models are all run on an H2O cluster which is started by the package during the initialization of the training step.  If a cluster has been previously initialized on the system being used 
 , the package will connect to it. Parameters for the training models can be specified by the user using the configuration files. 

If the user desires, the package can run a second stage of training, where the first stage model is used to generate predictions for all of the training data, which are then spatially weighted to serve as representations of nearby pollution as additional inputs for the second round of training. When indicated, this is done to allow us to improve our predictions by bringing in information from nearby monitors or grid points. 

\emph{Predictions}
Predictions are generated based on the trained model, but before these predictions are generated the data being used for prediction must go through the same preparation steps that were used to clean and prepare the training data. All of the inputs for those steps are saved during the preparation process, and the prediction step retrieves those and uses them to repeat the imputation, transformation, and normalization calculations (if they were done). 

The predictions based on the ensemble model, as well as the individual models used to generate it, are all saved during the prediction step. The prediction step recreates the the outputs from each individual model, then uses them as inputs in a GAM model, producing the final ensemble predictions. If the two stage modelling process is being used, the prediction code will handle the spatial weighting and output the predictions from the second stage model as well. 

\subsection{Flexibility}
In order to maximize flexibility while at the same time creating easy to use functions and tools, we chose to implement most algorithmic specifications using configuration files. These were implemented in the YAML (Yet Another Mark-up Language) format. YAML is a mark-up language that is both human and machine readable. It is a series of fields and values, with the field name existing at the highest level of the document and values following a colon placed after the field name. There are a number of libraries within R that ease the process of reading YAML files, and allow for strings, floats, and Boolean values and vectors to be easily written and read. 

The primary configuration file in the R package is called config.yml. It can be generated using the gen\textunderscore config() function. The config file will be placed in the current working directory with established defaults in place for the settings. The file can either be edited directly by the user, or take a list containing field names and associated values that will overwrite all defaults as input. The fields in this configuration primarily specify the paths to data inputs, but also have some some values specifying larger work-flow considerations, such as whether or not the imputation algorithm should be run, as well as the specific models to be used in the ensemble model.

In addition to the primary configuration file, there is also the option of using supplementary configuration files with each model. These allow for the specification of the parameters of each individual model being passed into the final ensemble. Each individual model has a corresponding configuration file, where each field is a parameter used by the H2O function which is called by the package. This feature allows users to specify model parameters without having to directly change the R code being run. While the executed functions remain the same, the user can adjust the layout of hidden layers in the neural net, the number of trees in the random forest, and the number of folds used for cross validation by changing a single value in the configuratoin file.

\begin{figure}
    \centering
    \includegraphics[width=0.5\textwidth]{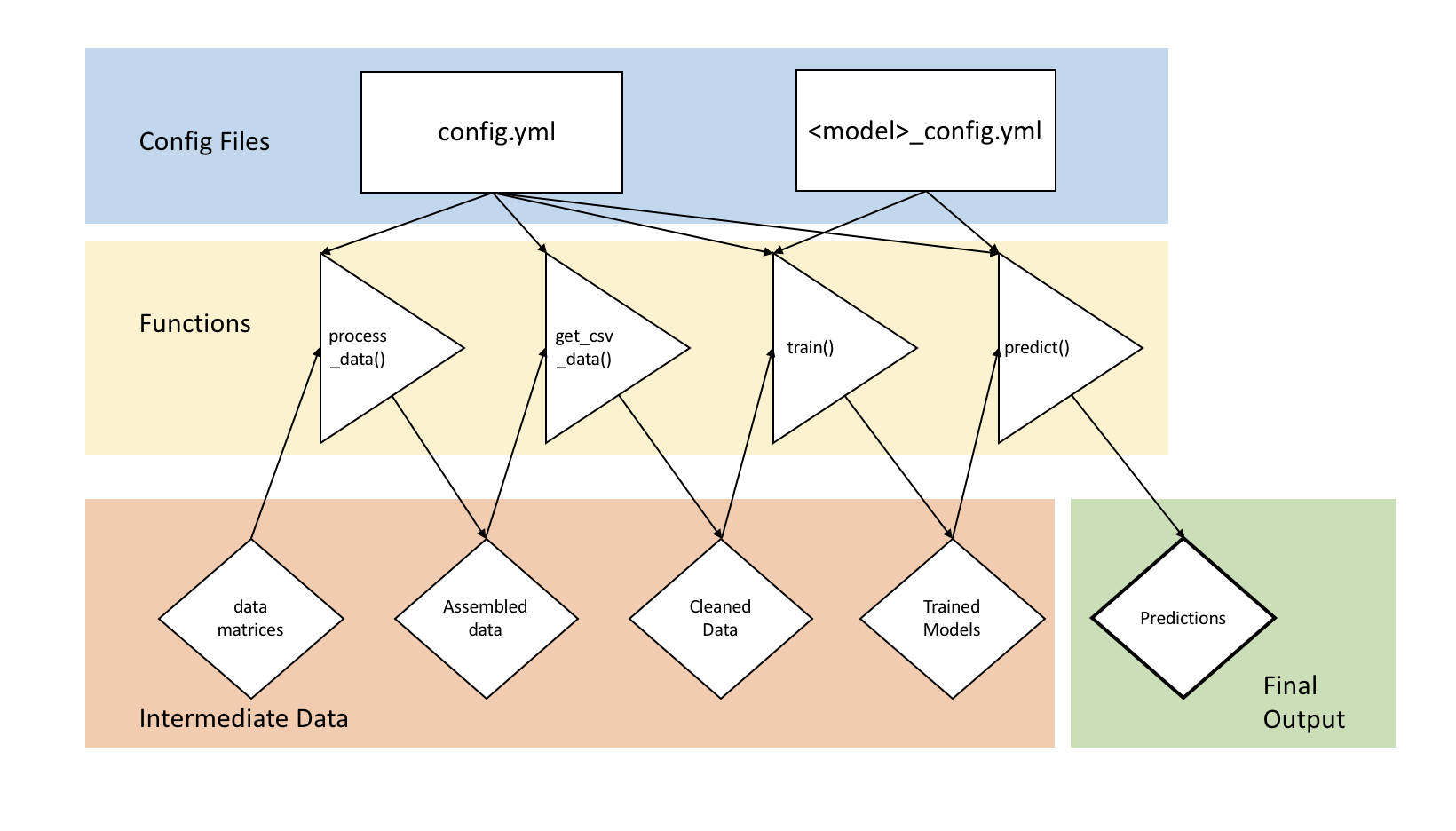}
    \caption{The configuration files and data used and produced during each step of the work flow}
    \label{fig:my_label}
\end{figure}

\subsection{User Experience}

The mix of technical and statistical problems that arise due to most machine learning models requiring adjustment of multiple tuning parameters, compared to the optimized algorithms of more traditional statistical models,
often lead to inefficient processing, misspecified models, or uncaught errors. Given the size of input data, and the time it takes to test determine the proper settings for the tuning parameters associated with each model, we felt it was important to both ease the process of model tuning and provide reasonable default settings so that a fair model could be constructed with minimal effort on the part of the user. As such, our goal is designing the package was to abstract away and automate as much of the complexity as we feasibly could. We also sought to ensure that code written using the package would still allow users unfamiliar with the h2o implementation of the model they would like to implement 
 to have a reasonable idea of what was being done. 

To implement these principles, we took advantage of the configuration file structure developed relatively few, clearly named, functions that implement the entire work flow from data assembly to prediction. For example, the entire work flow consists of the following lines of code:

\begin{verbatim}
    library(airpred)
    read_csv_data()
    train()
\end{verbatim}

We developed a specific structure for the configuration file, and each configuration files must have all of the appropriate fields filled out in order to run. An example of a primary configuration file (config.yml) is shown here:

\begin{verbatim}
    csv_path: input_data/pm25_jan_2012_data.csv
    ...
    normalize: TRUE
    transform: TRUE
    impute: TRUE
    imputation_models: imputation_models
    mid_process_data: mid_process_data
    training_data: mid_process_data/prepped.RDS
    training_output: training_output
    ...
    two_stage: FALSE
    models:
    - nn
    - forest
    - gradboost
\end{verbatim}

This configuration file specifies 1) the location of the input data which is stored as an unprocessed csv, 2) whether or not the data cleaning should include the normalization, transformation, and imputation steps, 3) where to store the imputation models, 4) where to store the partially processed data sets that are produced after each step during the data cleaning process for both documentation and debugging purposes 
), 5) where the processed data to be used for training is stored, and 6) where to store all files and objects generated through the training process (training\textunderscore output).  The models field uses YAML vector notation to store the list of selected models to be combined in the ensemble step. The three models being run in this specific example are the neural net, random forest, and gradient boosting models.

 All paths are specified as relative paths, meaning that we can use this interface to create a common folder structure and easily transport the whole platform between computer systems. Additionally, assuming that the data is prepared using the functions referenced above, by default the outputs will always be placed in the mid\textunderscore process\textunderscore data directory with the name "prepped.RDS". 

\section{Discussion}

We implemented and developed a complete work-flow required to predict various pollutant concentrations in R. The developed airpred R package has conferred a number of benefits to our group and makes public a tool that we hope will enable more researchers to improve their processes in similar fields of research. 

This package was developed to be usable on multiple systems, a fact we have confirmed through our tests on both cloud and cluster computing systems. This flexibility increases the number of potential applications for the platform, as well as allowing instances of models built using the package to be easily transported between systems.

The development of this package has significantly improved the ability of researchers within our group to develop their air pollution models. By wrapping the functionality of H2O, we can utilize a system that is significantly faster\cite{pafka_benchmarking_2015} than R while still remaining in the R framework. The configuration files also have improved our documentation processes as we now have a record of all models and their specifications. This can allow for a record of past models to be maintained, or allow for multiple models to be developed in parallel to each other. A consistent, portable file structure can be implemented so that creating a new model only requires copying a small number of folders and changing the configuration files.  This platform has been developed in a way that will easily allow the code to be shared with researchers working on similar problems. The code is currently hosted on GitHub, and is available to be distributed in an open source way.

An additional advantage of this package is that it was developed using standard software development best practices. Best practices that define quality code have been established for quite a while. Unfortunately, these practices are not always followed in the development of scientific software. Frequently, software doesn't make use of functions and tools for documentation to ensure that code can be easily maintained. Modularity in data structures and processes significantly improves the maintainability of any piece of code in addition to being a standard software engineering practice. 

Even though the code was developed to model PM$_{2.5}$ and has been illustrated in this paper using examples drawn from PM$_{2.5}$ modelling, there is nothing in the package that ties the code to PM$_{2.5}$ modelling. Other pollutants that this has been tested with include ozone and NO$_2$. In theory, this framework could be used with additional pollutants, such as PM$_{2.5}$ speciation, or black carbon.


There are some limitations to the work that leave room for further development. First of which is that we require the user to provide the data prepared to be in one of the structures that the package has been designed to process. While we do specify the form that the data needs to be in in order to be processed, the code can not currently handle deviations from this shape, nor does it provide routines to aid users in processing their data into this form.
The package can be extended to incorporate more complex data cleaning algorithms, as well as incorporating basic Geographic Information Systems algorithms, potentially with calls to Postgis. 

An additional limitation is that there are only a limited number of models currently implemented within the airpred framework. Additional models that are implemented in H2O could be added with minimal development. However, the reliance on H2O is also a limitation. Future development could allow airpred to access functionality from other Big Data tools, such as Spark or TensorFlow.


Ultimately, the development of this package will facilitate the ability of researchers looking to use or develop models that incorporate data that vary over space and time, especially those that implement predictions of various forms of pollution. By reducing the amount of time necessary for researchers to develop and ensure functionality of the IT elements and the data cleaning processes, as well as providing easier interfaces along with functions that speed computations, the package allows researchers to easily train new models. 

\acknowledgments{
The authors wish to thank the Harvard Research Computing Environment, the Harvard Center for Geographic Analysis, and the Massachusetts Open Cloud who provided computational infrastructure and support for this work. We also want to thank Joel Schwarz for his intellectual guidance. Research described in this article was conducted under contract to the Health Effects Institute (HEI), an organization jointly funded by the United States Environmental Protection Agency
(EPA) (Assistance Award No. R-83467701) and certain motor vehicle and engine manufacturers.
The contents of this article do not necessarily reflect the views of HEI, or its sponsors, nor do they
necessarily reflect the views and policies of the EPA or motor vehicle and engine manufacturers. 
}

\bibliographystyle{abbrv-doi}

\bibliography{airpred}
\end{document}